\documentclass[10pt,twocolumn,letterpaper]{article}

\usepackage{iccv}
\usepackage{times}
\usepackage{epsfig}
\usepackage{graphicx}
\usepackage{color}
\usepackage{caption}
\usepackage{amsmath}
\usepackage{amssymb}
\usepackage{bm}
\usepackage{booktabs}
\usepackage{lib}
\usepackage{algorithm}
\usepackage{algpseudocode}

\algnewcommand{\Initialize}[1]{%
  \State \textbf{Initialize:}
  \Statex \parbox[t]{\linewidth}{\raggedright #1}
}

\usepackage[pagebackref=true,breaklinks=true,letterpaper=true,colorlinks,bookmarks=false]{hyperref}

\iccvfinalcopy 

\def\iccvPaperID{1772} 

\ificcvfinal\fi
\begin{document}

\title{Amodal Completion and Size Constancy in Natural Scenes}

\author{Abhishek Kar, Shubham Tulsiani, Jo\~{a}o Carreira and Jitendra Malik\\
University of California, Berkeley - Berkeley, CA 94720\\
{\tt\small \{akar,shubhtuls,carreira,malik\}@eecs.berkeley.edu}}

\maketitle

\begin{abstract}

We consider the problem of enriching current object detection systems with veridical object sizes and relative depth estimates from a single image. There are several technical challenges to this, such as occlusions, lack of calibration data and the scale ambiguity between object size and distance. These have not been addressed in full generality in previous work. Here we propose to tackle these issues by building upon advances in object recognition and using recently created large-scale datasets. We first introduce the task of amodal bounding box completion, which aims to infer the the full extent of the object instances in the image. We then propose a probabilistic framework for learning category-specific object size distributions from available annotations and leverage these in conjunction with amodal completions to infer veridical sizes of objects in novel images. Finally, we introduce a focal length prediction approach that exploits scene recognition to overcome inherent scale ambiguities and demonstrate qualitative results on challenging real-world scenes.
\end{abstract}
\section{Introduction}


Consider \figref{fig1}. Humans can effortlessly perceive two chairs of roughly the same height and tell that one is much closer than the other, though still further away than the person, who is taller than the chairs. Compare this to what a state-of-the-art object detector tells us about the image: that there are two chairs, 120 and 40 pixels tall, and one person with 200 pixels from top to bottom. How can we enable computer vision systems to move beyond this crude 2D representation and allow them to capture richer models of their environments, such as those that humans take for granted?

The 3D world is a lot more structured than it looks like from the retina (or from a camera sensor), where objects jump around with each saccade and grow and shrink as we move closer or farther from them. We do not perceive any of this because our brains have learned priors about how visual inputs correlate with the underlying environment, and this allows us to directly access realistic and rich models of scenes. The priors we use can be categorized as being related to either \textit{geometry} or \textit{familiarity}.

\begin{figure}[t!]
  \centering
  \includegraphics[width=0.48\textwidth]{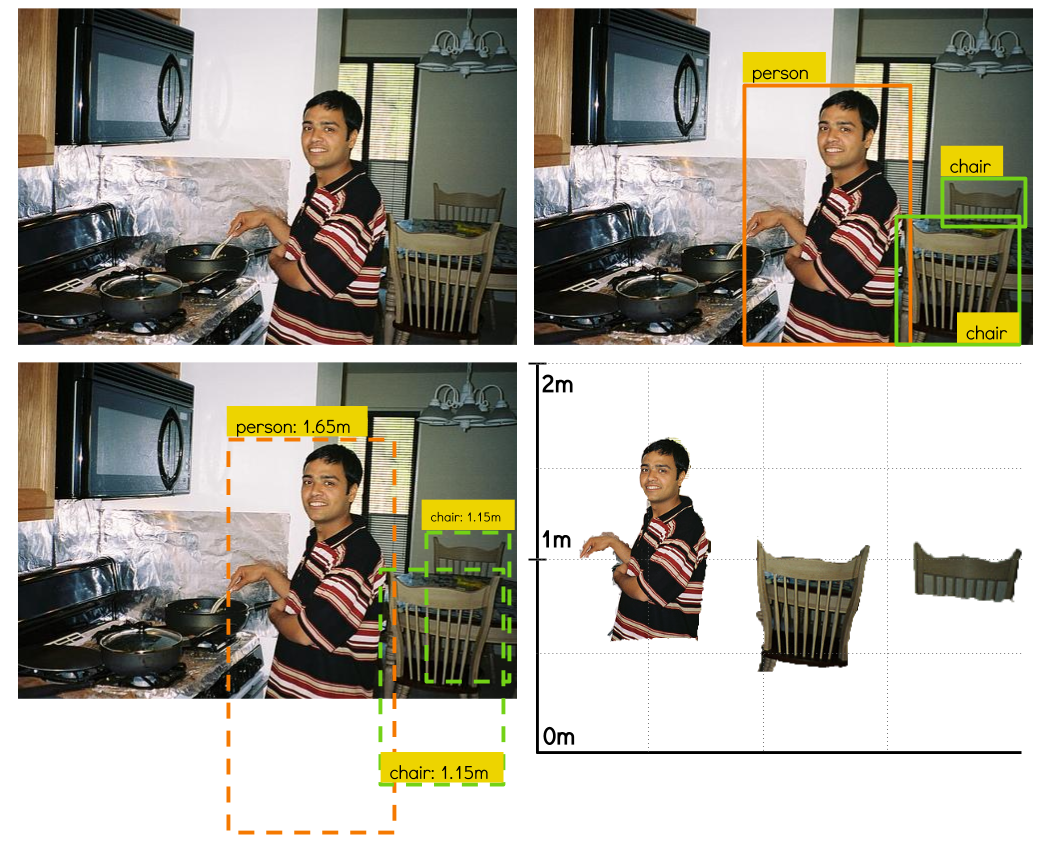} 
  \caption{\figlabel{fig1} Perceiving the veridical size of objects in realistic scenes, from a single image, requires disentangling size and depth, being able to compensate for occlusions and to determine intrinsic camera parameters. We tackle all three of these problems, leveraging recent developments in object recognition and large annotated object and scene datasets.}
\end{figure}

Image projection properties, such as the fact that the distance of an object from the camera dictates apparent size and that parallel lines in the scene vanish in the image, provide useful signal for perceiving structure. Familiarity cues are complementary and impose expectations on individual objects and configurations -- we expect most objects to be supported by another surface and we have the notion of \textit{familiar size} -- similar objects are of similar sizes. In this work, we exploit geometry and familiarity cues and develop a framework to build richer models of the visual input than those given by current computer vision systems, which are still largely confined to the 2D image plane.

The notion that certain geometrical cues can aid perception has been known since the time of Euclid - the points in the image where objects touch the ground together with their perceived heights allows inference of real world object size ratios \cite{burton45}. Familiarity cues, on the other hand must be learned, which can be done using available annotations and building upon rapid recent progress in object recognition, more robustly harnessed to explain novel images. Similar ideas have been proposed by Hoiem \etal~\cite{hoiem2008putting,Hoiem:book} and Gupta \etal ~\cite{gupta2010blocks} who studied the interaction between object detection and scene layout estimation and showed that, by reasoning over object sizes within their 3D environment, as opposed to within the image, one could perform better object detection. Lalonde \etal~\cite{lalonde2007photo} and Russell \etal~\cite{russell2009building} also tackled a problem similar to operationalizing size constancy and inferred object sizes of annotated objects. These works, while sharing similar goals to ours, were limited in their scope as they assumed fully visible instances - object recognition technology at the time being a limiting factor. In this paper, we aim for veridical size estimation in more realistic settings -- where occlusions are the rule rather than the exception. Occlusions present a significant technical challenge as they break down a number of assumptions(\eg in 
\figref{fig1} not modeling occlusions would yield an incorrect estimate of the relative depths of the two chairs shown).

To overcome these challenges, we first introduce amodal completion. This is a very well studied ability of human perception, primarily in the context of amodal edge perception \cite{kanizsa1979organization}, building on theories of \textit{good continuation} \cite{shipley2001fragments}. In the context of objects, amodal completion manifests itself as inference of the complete shape of the object despite visual evidence for only parts of it \cite{breckon2005amodal}.
In \secref{amodalCompletion}, we tackle the amodal completion task and frame it as a recognition problem, formalized as  predicting the full extent of object bounding boxes in an image, as opposed to only the visible extent. We build amodal extent predictors based on convolutional neural networks which we train on the challenging PASCAL VOC dataset. In \secref{sizeconstancy}, we propose a formulation that, leveraging amodally completed objects, can disentangle relative object sizes and object distances to the camera. This geometric reasoning allows us only to infer distances for objects up to a scaling ambiguity in each image. To overcome this ambiguity, we show in \secref{sceneFocal} that it is possible to leverage statistical dependencies between scenes and intrinsic camera parameters, and learn to predict focal lengths of scenes from large scale scene datasets. Finally, we present qualitative  results exhibiting veridical size estimation in complex scenes.

\section{Amodal Completion}
\seclabel{amodalCompletion}
``Almost \textit{nothing} is visible in its entirety, yet almost \textit{everything} is perceived as a whole and complete" \cite{palmer1999vision}.
Classic computer vision approaches have traditionally been impoverished by trying to explain just what we see in an image. For years, standard benchmarks have focused on explaining the visible evidence in the image - not the world behind it. For example, the well-studied task of predicting the bounding box around the visible pixels of an object has been the goal of current object detection systems. As humans, not only can we perceive the visible parts of the chair depicted in \figref{fig1}, we can confidently infer the full extent of the actual chair.

This representation of objects, that humans can effortlessly perceive, is significantly richer than what current systems are capable of inferring. We take a step forward towards achieving similar levels of understanding by attacking the task of perceiving the actual extent of the object, which we denote as \textit{amodal completion}. The amodal representation of objects enables us to leverage additional scene information such as support relationships, occlusion orderings etc. For example, given the amodal and visible extents of two neighboring objects in the image, one can figure out if one is occluded by the other. Explicitly modeling amodal representations also allow us to implicitly model occlusions patterns rather than trying to ``explain them away" while detecting objects. As described in \secref{sizeconstancy}, we can use these representations to infer real world object sizes and their relative depths just from images.

The primary focus of object recognition systems \cite{girshick2013rich,felzens_latent_pami10} has been to localize and identify objects, despite occlusions, which are usually handled as noise. Several recently proposed recognition systems do explicitly model occlusion patterns along with detections and provide a mechanism for obtaining amodal extent of the object \cite{ghiasi2014parsing, xiang_cvpr15, zia2014towards}. However, these approaches have been shown to work only on specific categories and rely on available shape models or depth inputs, for learning to reason over occlusions. In contrast, we aim to provide a generic framework that is not limited by these restrictions. Our proposed framework is described below.


\paragraph{Formulation:} Given a candidate visible bounding box, we tackle the task of amodal completion -- the input to our system is some modal bounding box (\eg obtained via a detection system) and we aim to predict the amodal extent for the object. We frame this task as predicting the amodal bounding box, which is defined as  the bounding box of an object in the image plane if the object were completely visible, \ie if inter-object occlusions and truncations were absent. The problem of amodal box prediction can naturally be formulated as a regression task - given a noisy modal bounding box of an object we regress to its amodal bounding box coordinates. The amodal prediction system is implicitly tasked with learning common occlusion/truncation patterns and their effects on visible object size. It can subsequently infer the correct amodal coordinates using the previously learned underlying visual structure corresponding to occlusion patterns. For example, the learner can figure out that chairs are normally vertically occluded by tables and that it should extend the bounding box vertically to predict the full extent of the chair.

Let $b = (x,y,w,h)$ be a candidate visible (or modal) bounding box our amodal prediction system receives ($(x,y)$ are the co-ordinates of the top-left corner and $(w,h)$ are the width and height of the box respectively) and $b^* = (x^*,y^*,w^*,h^*)$ be the amodal bounding box of the corresponding object, our regression targets are $(\frac{x-x^*}{w},\frac{y-y^*}{h},\frac{(x+w)-(x^*+w^*)}{w},\frac{h-h^*}{h})$. Our choice of targets is inspired by the fact that for the $y$ dimension, the height and bottom of the box are the parameters we actually care about (see \secref{sizeconstancy}) whereas along the $x$ dimension the left co-ordinate is not necessarily more important than the right.

\paragraph{Learning:}

We use a Convolutional Neural Network (CNN) \cite{neocognitron,LeCun1989} based framework to predict the co-ordinates of the amodal bounding box. THe hypothesis is that the amodal prediction task can be reliably addressed given just the image corresponding to the visible object region -- seeing the left of a car is sufficient to unambiguously infer the full extent without significantly leveraging context. Based on this observation, we extract from input image $I$, the region corresponding to the detection box $b$ and train the CNN using targets derived as above from the amodal box $b^*$. We impose an $L_2$ penalty on the targets and regress from the extracted CNN image features to the targets. We initialize our model using the AlexNet \cite{krizhevsky2012imagenet} CNN pretrained for Imagenet \cite{imagenet_cvpr09} classification and then finetune the model specific to our task using backpropagation. Training is carried out with jittered instances of the ground truth bounding box to enable generalization from noisy settings such as detection boxes and also serve as data augmentation.

We train two variants of the above network - class-specific and class agnostic. Both these systems comprise of 5 convolutional layers followed by 3 fully-connected layers. The class-specific network has separate outputs in the last layers for different classes and is trained with positive examples from a specific class whereas the class agnostic network has a single set of outputs across all classes. Intuitively, the class-specific network learns to leverage occlusion patterns specific to a particular class (\eg chair occluded by a table) whereas the class agnostic network tries to learn occlusion patterns common across classes. Another argument for a class agnostic approach is that it is unreasonable to expect annotated amodal bounding box data for a large number of categories. A two-stage system, where we first predict the visible bounding box candidates and then regress from them to amodal boxes, enables leveraging these class agnostic systems to generalize to more categories. As we demonstrate in \secref{sizeconstancy}, this class agnostic network can be applied to novel object categories to learn object sizes.

\begin{figure}
  \centering
  \includegraphics[width=0.48\textwidth]{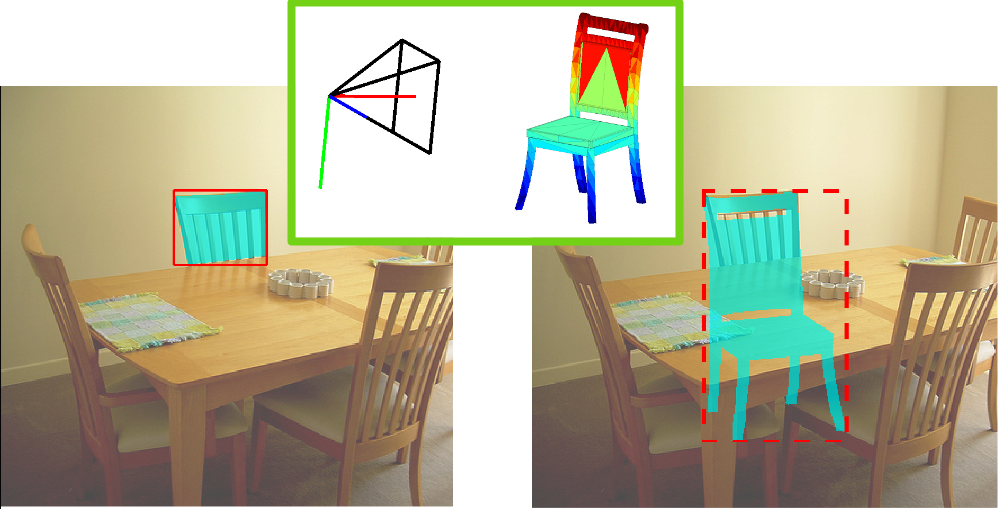}
  \caption{\figlabel{amodal} Generating amodal bounding boxes for instances in PASCAL VOC. We use the 3D models aligned to images from the PASCAL 3D+ \cite{pascal3d} and render them with their annotated 3D pose to obtain binary masks. We then use the tightest fitting bounding box around the mask as our ground truth amodal bounding box.}
\end{figure}

\paragraph{Dataset:} For the purpose of amodal bounding box prediction, we need annotations for amodal bounding boxes (unlike visible bounding box annotations present in all standard detection datasets). We use the PASCAL 3D+~\cite{pascal3d} dataset which has approximate 3D models aligned to 12 rigid categories on PASCAL VOC~\cite{pascal-voc-2012} to generate these amodal bounding box annotations. It also contains additional annotations for images from ImageNet \cite{imagenet_cvpr09} for each of these categories (~22k instances in total from ImageNet). For example, it has between 4 different models aligned to ``chair'' and 10 aligned to ``cars''. The different models primarily distinguish between subcategories (but might also be redundant). The 3D models in the dataset are first aligned coarsely to the object instances and then further refined using keypoint annotations. As a consequence, they correctly capture the amodal extent of the object and allow us to obtain amodal ground-truth.  We project the 3D model fitted per instance into the image, extract the binary mask of the projection and fit a tight bounding box around it which we treat as our amodal box (\figref{amodal}). We train our amodal box regressors on the detection training set of PASCAL VOC 2012 (\textit{det-train}) \textit{and} the additional images from ImageNet for these 12 categories which have 3D models aligned in PASCAL 3D+ and test on the detection validation set (\textit{det-val}) from the PASCAL VOC 2012 dataset.

\paragraph{Experiments:} We benchmark our amodal bounding box predictor under two settings - going from ground truth visible bounding boxes to amodal boxes and in a detection setting where we predict amodal bounding boxes from noisy detection boxes. We compare against the baseline of using the modal bounding box itself as the amodal bounding box (\textit{modal bbox}) which is in fact the correct prediction for all untruncated instances. \tableref{gtBboxTable} summarizes our experiments in the former setting where we predict amodal boxes from visible ground truth boxes on various subsets of the dataset and report the mean IoU of our predicted amodal boxes with the ground truth amodal boxes generated from PASCAL 3D+. As expected, we obtain the greatest boost over the baseline for truncated instances. Interestingly, the class agnostic network performs as well the class specific one signaling that occlusion patterns span across classes and one can leverage these similarities to train a generic amodal box regressor. 

To test our amodal box predictor in a noisy setting, we apply it on bounding boxes predicted by the RCNN\cite{girshick2013rich} system from Girshick \etal. We assume a detection be correct if the RCNN bounding box has an IoU $> 0.5$ with the ground truth visible box \textit{and} the predicted amodal bounding box also has an IoU $> 0.5$ with the ground truth amodal box. We calculate the average precision for each class under the above definition of a ``correct'' detection and call it the Amodal $AP$ (or $AP^{am}$). \tableref{detectionTable} presents our $AP^{am}$ results on VOC2012 \textit{det-val}. As we can see again, the class agnostic and class specific systems perform very similarly. The notable improvement is only in a few classes (\eg diningtable and boat) where truncated/occluded instances dominate. Note that we do not rescore the RCNN detections using our amodal predictor and thus our performance is bounded by the detector performance. Moreover, the instances detected correctly by the detector tend to be cleaner ones and thus the baseline (\textit{modal bbox}) of using the detector box output as the amodal box also does reasonably well. Our RCNN detector is based on the VGG16 \cite{simonyan2014very} architecture and has a mean $AP$ of $57.0$ on the 12 rigid categories we consider.

\begin{table}[htb!]
\centering
\begin{tabular}{ccccc}
\toprule
& \textbf{all} & \textbf{trun/occ} & \textbf{trunc} & \textbf{occ} \tabularnewline
\midrule

\textbf{modal bbox} & 0.66 & 0.59 & 0.52 & 0.64 \tabularnewline
\textbf{class specific} & 0.68 & 0.62 & 0.57 & 0.65 \tabularnewline
\textbf{class agnostic} & 0.68 & 0.62 & 0.56 & 0.65 \tabularnewline

\bottomrule
\end{tabular}
\caption{Mean IoU of amodal boxes predicted from the visible bounding box on various subsets of the validation set in PASCAL VOC. Here \textit{occ} and \textit{trunc} refer to occluded and truncated instances respectively. The class specific and class agnostic methods refer to our variations of the training the amodal box regressors (see text for details) and modal bbox refers to the baseline of using the visible/modal bounding box itself as the predicted amodal bounding box.}
\tablelabel{gtBboxTable}
\end{table}

\begin{table*}[htb!]
\centering
\begin{tabular}{ccccccccccccc|c}
\toprule
& \textbf{aero} & \textbf{bike} & \textbf{boat} & \textbf{bottle} & \textbf{bus} & \textbf{car} & \textbf{chair} & \textbf{table} & \textbf{mbike} & \textbf{sofa} & \textbf{train} & \textbf{tv} & \textbf{mean}\tabularnewline
\midrule

\textbf{modal bbox} & 70.0 & 66.2 & 23.9 & 35.1 & 76.4 & 57.7 & 28.9 & 24.2 & 68.3 & 45.8 & 58.1 & 59.6 & 51.2 \tabularnewline
\textbf{class specific} & 69.5 & 67.2 & \textbf{26.9} & 36.0 & \textbf{77.0} & \textbf{61.4} & \textbf{31.4} & 29.2 & \textbf{69.0} & \textbf{49.4} & \textbf{59.3} & 59.5 & \textbf{53.0} \tabularnewline
\textbf{class agnostic} & \textbf{70.0} & \textbf{67.5} & 26.8 & \textbf{36.3} & 76.8 & 61.3 & 31.1 & \textbf{30.9} & 68.9 & 48.4 & 58.6 & \textbf{59.6} & \textbf{53.0} \tabularnewline
\bottomrule
\end{tabular}
\caption{$AP^{am}$ for our amodal bounding box predictors on VOC 2012 \textit{det-val}. $AP^{am}$ is defined as the average precision when a detection is assumed to be correct only when both the modal and amodal bounding boxes have IoU $> 0.5$ with their corresponding ground truths.}
\tablelabel{detectionTable}
\end{table*}

Armed with amodal bounding boxes, we now show how we tackle the problem of inferring real world object sizes from images.
\section{Untangling Size and Depth}
\seclabel{sizeconstancy}
Monocular cues for depth perception have been well-studied in psychology literature and there are two very important cues which emerge that tie object size and depth -  namely familiar size and relative size. Familiar size is governed by the fact that the visual angle subtended by an object decreases with distance from the observer and prior knowledge about the actual size of the object can be leveraged to obtain absolute depth of the object in the scene. Relative size, on the other hand, helps in explaining relative depths and sizes of objects - if we know that two objects are of similar sizes in the real world, the smaller object in the image appears farther. Another simple cue for depth perception arises due to perspective projection - an object further in the world appears higher on the image plane. Leveraging these three cues, we show that one can estimate real world object sizes from just images. In addition to object sizes, we also estimate a coarse viewpoint for each image in the form of the horizon and camera height. 

The main idea behind the algorithm is to exploit pairwise size relationships between instances of different object classes in images. As we will show below, given support points of objects on the ground and some rough estimate of object sizes, one can estimate the camera height and horizon position in the image - and as a result relative object depths. And in turn, given object heights in the image and relative depths, one can figure out the real world object scale ratios. Finally, exploiting these pairwise size evidences across images, we solve for absolute real world sizes (upto a common scale factor or the metric scale factor). Note that we use size and height interchangeably here as our notion of object size here actually refers to the object height.
\begin{figure}
  \centering
  \includegraphics[width=0.48\textwidth]{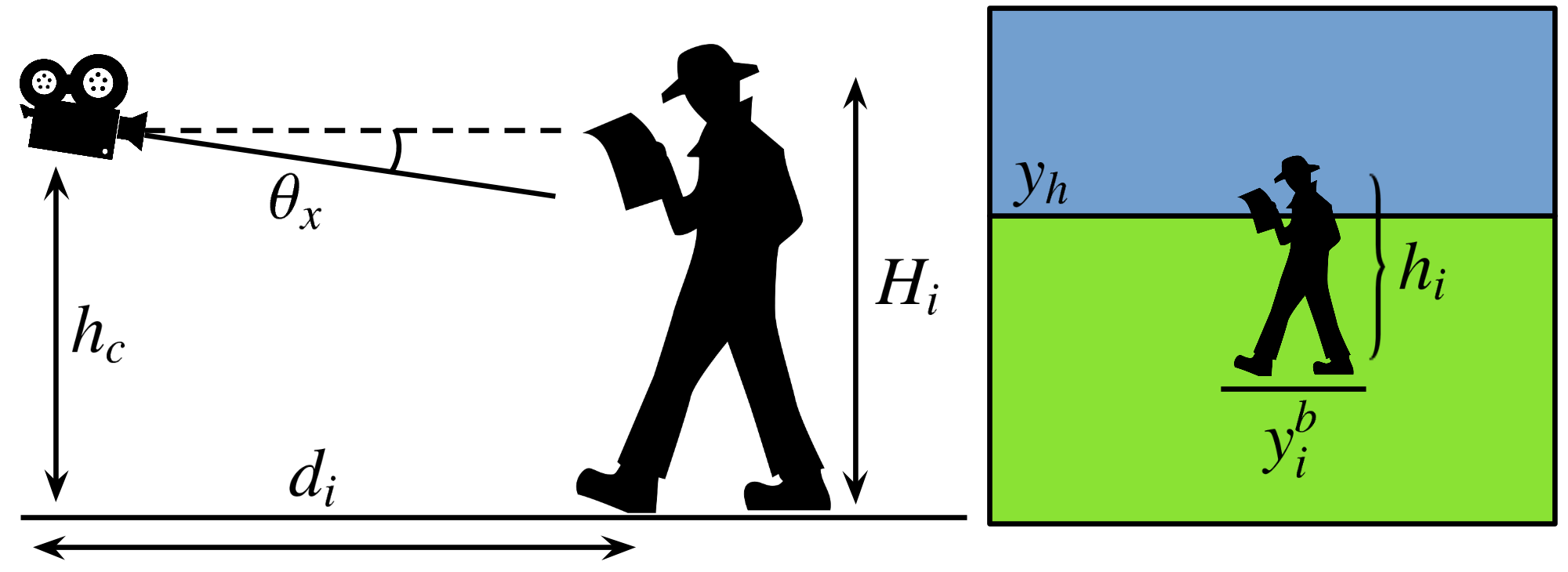}
  \caption{\figlabel{perspective} Toy example illustrating our camera model and parameters. Please refer to the text for detailed explanations. }
\end{figure}

\paragraph{Camera Model:} We use a simplified perspective camera model similar to Hoiem \etal~\cite{hoiem2008putting}. Let $f$ be the focal length of the camera, $\theta_x$ the camera tilt angle along the x-axis, $h_c$ the height of the camera, $y_h$ be the horizon position in the image, $y_i^b$ be the ground support point for the $i^{th}$ object in the image and $d_i$ be the distance of the $i^{th}$ object from the camera along the camera axis ($z$ axis). We assume that the images have been corrected for camera roll and all pixel co-ordinates are with respect to the optical center (assumed to be center of the image). \figref{perspective} provides a toy illustration of our model and parameters.

Assuming that the world frame is centered at the camera with its $y$ axis aligned with the ground, the projection of a world point $\mathbf{X} = (X_w,Y_w,Z_w)$ in the image in homogeneous co-ordinates is given by:
\bes 
\begin{bmatrix}
x \\
y \\
1
\end{bmatrix} = \frac{1}{Z_w}
\begin{bmatrix}
f & 0 & 0\\
0 & f & 0\\
0 & 0 & 1
\end{bmatrix}
\begin{bmatrix}
1 & 0 & 0 & 0 \\
0 & \cos\theta_x & \sin\theta_x & 0 \\
0 & -\sin\theta_x & \cos\theta_x & 0 \\
\end{bmatrix}
\begin{bmatrix}
X_w \\
Y_w \\
Z_w \\
1
\end{bmatrix}
\ees
For a world point corresponding to the ground contact point of object $i$, given by $(X_w,-h,d_i)$, its corresponding $y$ co-ordinate in the image $y_i^b$ is given by:
$
y_i^b = f\frac{(-h_c/d_i + \tan\theta_x)}{1+(h_c/d_i)\tan\theta_x}
$
Under the assumption of the tilt angle being small ($\tan\theta_x \approx \theta_x$) and height of the camera being not too large compared to object distance ($h\theta_x \ll d_i$), our approximation is 
\be
\eqlabel{groundEq}
y_i^b = -\frac{fh_c}{d_i} + f\theta_x
\ee
Here $f\theta_x$ corresponds to the position of the horizon ($y_h$) in the image. Repeating the above calculation for the topmost point of the object and subtracting from \eqref{groundEq}, we obtain
\be 
\eqlabel{sizeEq}
h_i = \frac{fH_i}{d_i}
\ee
where $h_i$ refers to the height of the object in the image and $H_i$ is the real world height of the object.
Our model makes some simplifying assumptions about the scene namely, objects are assumed to rest on the same horizontal surface (here, the ground) and camera tilt is assumed to be small. We observe that for the purpose of size inference, these assumptions turn out to be reasonable and allow us to estimate heights of objects fairly robustly.

\begin{algorithm}[h]
\caption{Object Size Estimation}
\begin{algorithmic}
\Initialize{Initial size estimates $\mathbf{H}$ and cluster assignments}
\While {not converged}
\ForAll{images $k \in $ Dataset }
\State $(h_c,y_h) \gets $ SolveLeastSquares($y_b,h,\mathbf{H}$)
\ForAll{pairs $(i,j)$ of objects in $k$}
\State $\frac{H_i}{H_j} \gets \frac{h_i}{h_j}\frac{y_j^b-y_h}{y_i^b-y_h}$\Comment{$(1)$}
\EndFor
\EndFor
\State $\log \mathbf{H} \gets$ least squares with pairwise constraints ($1$)
\State GMM cluster log scales ($\log \mathbf{H}$)
\State Reassign objects to clusters
\EndWhile
\end{algorithmic}
\label{alg:sizeEstimation}
\end{algorithm}

\paragraph{Inferring Object Sizes:} 
The important observation here is that the sizes of objects in an object category are not completely random - they potentially follow a multimodal distribution. For example, different subcategories of boats may represent the different modes of the size distribution. Given some initial sizes and size cluster estimates, our algorithm for size estimation (Algorithm \ref{alg:sizeEstimation}) works by estimating the horizon and camera height per image (by solving a least squares problem using \eqref{groundEq} and \eqref{sizeEq} for all the objects in an image). With the horizon and height estimated per image, we obtain pairwise height ratios $\frac{H_i}{H_j} = \frac{h_i}{h_j}\frac{y_j^b-y_h}{y_i^b-y_h}$ for each pair of objects in an image. We obtain multiple such hypotheses across the dataset which we use to solve a least squares problem for $\log \mathbf{H}$ - the log height for each size cluster. Finally, we cluster the log sizes obtained in the previous step to obtain new size clusters and iterate. Note that $\mathbf{H}$ refers to the vector with heights of various classes and $H_i$ refers to the real world size of the $i^{th}$ object.

This particular model is equivalent to solving a latent variable model where the latent variables are the cluster memberships of the instances, the estimated variables are heights corresponding to the size clusters and the horizon and camera height for each image. The loss function we try to minimize is the mean squared error between the ground contact point predicted by the model and the amodal bounding box. Finally, the log of the object heights are assumed to be a Gaussian mixture. This final assumption ties in elegantly with psychophysics studies which have found that our mental representation of object size (referred to as assumed size~\cite{ittelson1951size,baird1963retinal,epstein1963influence}) is proportional to the logarithm of the real world object size~\cite{konkle2011canonical}.

Our image evidences in the above procedure include the ground support points and heights for all the objects in the image. Note that amodal bounding boxes for objects provide us exactly this information. They account for occlusions and truncations and give us an estimate of the full extent of the object in the image. The above algorithm with occluded/truncated visible bounding boxes would fail miserably and we use our amodal bounding box predictor to first ``complete'' the bounding boxes for us before using our size inference algorithm to infer object heights.

\begin{figure}
  \centering
  \includegraphics[width=0.48\textwidth]{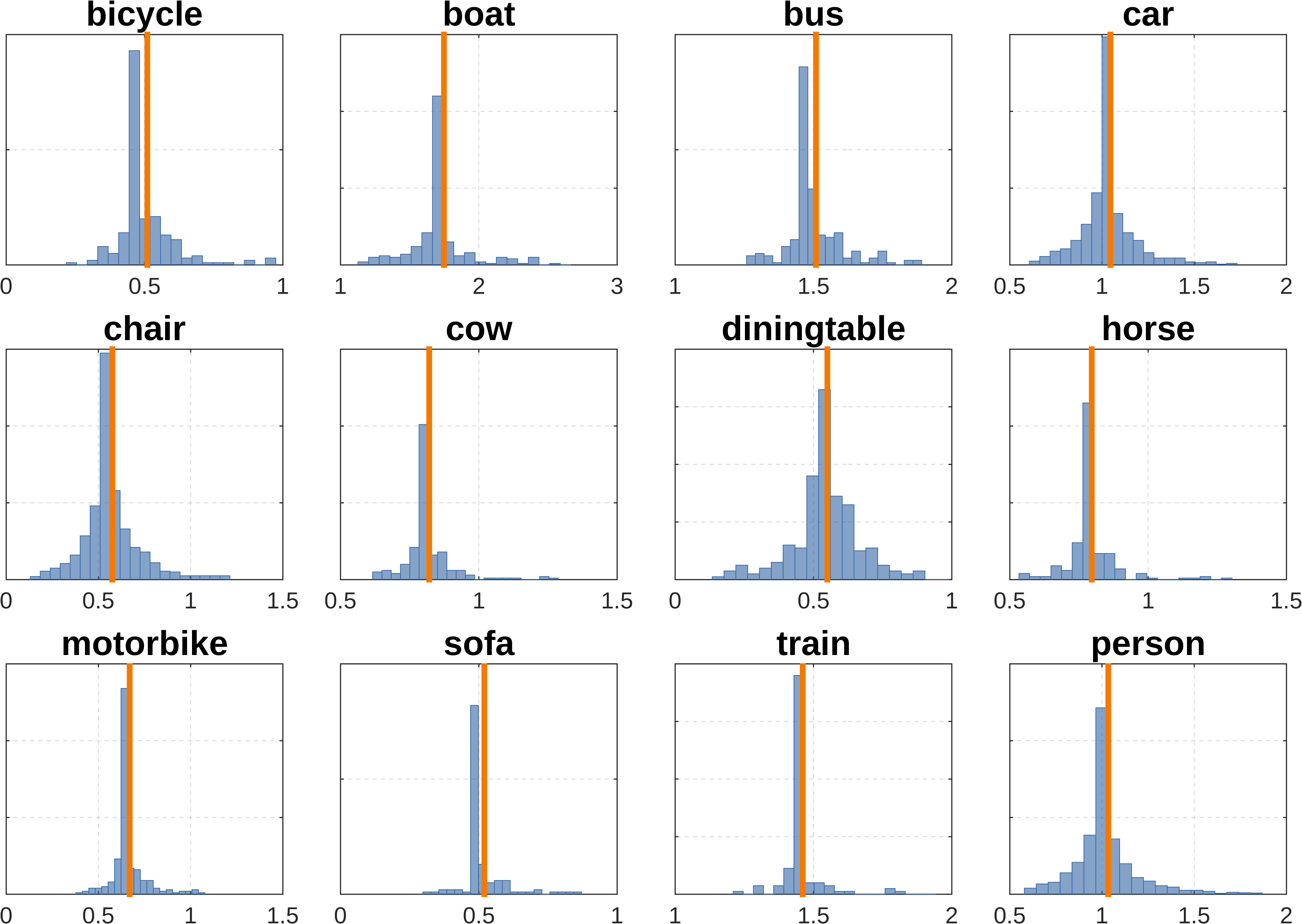}
  \caption{\figlabel{pascalSizes} Inferred log size distributions of 12 object categories on PASCAL VOC. We use our class agnostic amodal bounding box predictor to predict amodal boxes for all instances in VOC 2012 \textit{det-val} and use them with our object size estimation system to estimate size distributions for various categories. The plots above show distributions of the log size with the mean size being shown by the orange line.}
\end{figure}

\paragraph{Inferring Object Size Statistics on PASCAL VOC:} We used our size estimation system on PASCAL VOC to estimate size distributions of objects. First, we use our class agnostic amodal bounding box predictor on ground truth visible bounding boxes of all instances on VOC 2012 \textit{det-val} to ``upgrade'' them to amodal boxes. We initialize our system with a rough mean height for each object class obtained from internet sources (Wikipedia, databases of cars etc.) and run our size estimation algorithm on these predicted amodal boxes. \figref{pascalSizes} shows the distributions of log sizes of objects of various categories in PASCAL VOC. Most categories exhibit peaky distributions with classes such as ``boat'' and ``chair'' having longer tails owing to comparatively large intra class variation. Note that we experimented with using multiple size clusters per class for this experiment but the peaky, long tailed nature of these distributions meant that a single Gaussian capturing the log size distributions sufficed. In addition to inferring object sizes, we also infer the horizon position and height of the camera. The median height of the camera across the dataset was 1.4 metres (roughly the height at which people take images) and also exhibited a long tailed distribution (please refer to supplementary for details). Some examples of amodal bounding boxes estimated for all instances from visible bounding boxes and horizons are shown in \figref{resultFig}.

\begin{figure*}
  \centering
  \includegraphics[width=\textwidth]{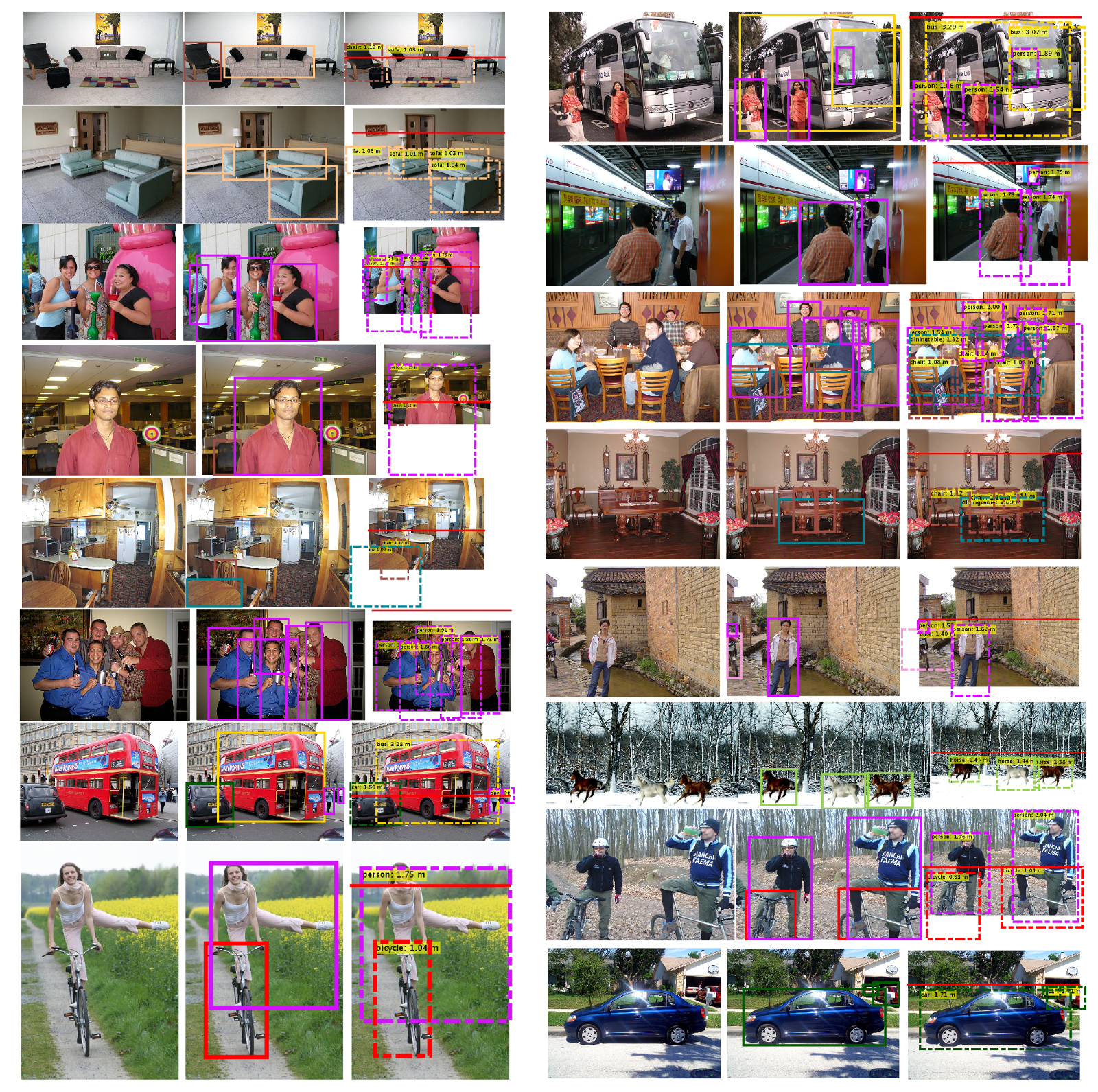} 

\caption{\figlabel{resultFig} Amodal bounding box prediction and size estimation results on images in PASCAL VOC. The solid rectangles represent the visible bounding boxes and the dotted lines are the predicted amodal bounding boxes with heights in meters. The horizontal red line denotes the estimated horizon position for the image.}
\end{figure*}
\section{Scenes and Focal Lengths}
\seclabel{sceneFocal}
The focal length of a camera defines its field of view and hence determines how much of a scene is captured in an image taken by the camera. It is an important calibration parameter for obtaining metric, as opposed to projective, measurements from images. The focal length is usually calibrated using multiple images of a known object \cite{zhang2000flexible}, such as a chessboard, or as part of bundle adjustment \cite{triggs2000bundle}, from multiple images of realistic scenes. It is one of the best studied sub-fields of computer vision -- e.g. see \cite{Hartley2004}. Well known existing approaches require a minimum set of vanishing lines  \cite{wang1991camera} or exploit Manhattan-world assumptions \cite{caprile1990using}. These techniques are very precise and elegant, but not generally applicable (e.g. beach or forest images, etc.).

We propose instead a learning approach that predicts focal length based on statistical dependencies between scene classes and fields of view. Given a same scene, images taken with large focal lengths will have fewer things in them than those captured with small focal lengths and this provides a cue for determining focal length. However certain scenes also have more things than others. This ambiguity can be resolved by training a predictor with many images of each scene class, taken with different focal lengths. 

Additionally, certain scenes tend to be pictured with preferred focal lengths. As an example, consider a scene class of ``pulpits". If a picture of a pulpit is taken with a short focal length, then the whole church will be visible and that image will not be tagged as a pulpit scene. In order for a pulpit to be dominant in a picture taken with a short focal length camera, then the photographer would have to be unnaturally close to it. 

\paragraph{Data:} We use the Places database \cite{zhou2014learning}, a large dataset that provides a dense sampling of scenes in natural images: it has $205$ scene classes, as diverse as \textit{swimming pool} and \textit{rope bridge}, and $2.5$ million images. We were able to scrape focal length metadata from EXIF tags of approximately $20$k examples, on average 100 per class, and split these into a training set having $15$k and a validation set of $5$k images. 

\paragraph{Learning:} We considered the problem of predicting the ratio of the focal length to the camera sensor width, which when multiplied by the size of the image in pixels gives the desired the focal length in pixels. We clustered the logarithm of this ratio into $10$ bins using k-means and formulated the prediction problem as classification, using a softmax loss. Images in the bin with highest and smallest focal length ratio are shown in \figref{focal_lengths}. We experimented finetuning different popular convolutional networks, including two trained on Imagenet classification -- AlexNet \cite{krizhevsky2012imagenet} and VGG-Deep16 \cite{simonyan2014very} -- and a network trained on the Places scenes -- the PlacesNet \cite{zhou2014learning}.

\paragraph{Results:} The results are shown in \tableref{focal_results} and suggest that focal length can indeed be predicted directly from images, at least approximately, and that pretraining on annotated scene class data makes a good match with this task. Our best model can predict correct focal length quite repeatably  among the top-three and top-five predictions. As baselines, we measure chance performance, and performance when picking the mode of the distribution on the training set -- the bin having most elements. Note that the bins are unbalanced (please refer to the supplementary material for the distribution of focal lengths across our dataset).

Note that our goal is not high precision of the type that is necessary for high-fidelity reconstruction; we aim for a coarse estimate of the focal length that can be robustly computed from natural images. Our results in this section are a first demonstration that this may be feasible.

\begin{table}
\centering
 \begin{tabular}{ l  c  c c}
\toprule
\textbf{ConvNet} & \textbf{top-1} & \textbf{top-3} & \textbf{top-5} \\
\midrule
Chance & 90.0 & 70.0 & 50.0 \\
Mode Selection & 60.2 & 26.4 & 8.7 \\
\midrule
AlexNet-Imagenet & 57.1  & 18.8 & 3.9 \\
VGG-Deep16-Imagenet & 55.8 & 15.9 & 3.3 \\
PlacesNet-Places & \textbf{54.3} & \textbf{15.3} & \textbf{3.1} \\
\bottomrule
  \end{tabular}
    \caption{\tablelabel{focal_results}Focal length misclassification rate (top-1, top-3 and top-5 predictions) of networks pretrained on object images from Imagenet and the Places dataset. Lower is better.}
\end{table}

\begin{figure}
  \centering
    \includegraphics[width=0.15\textwidth]{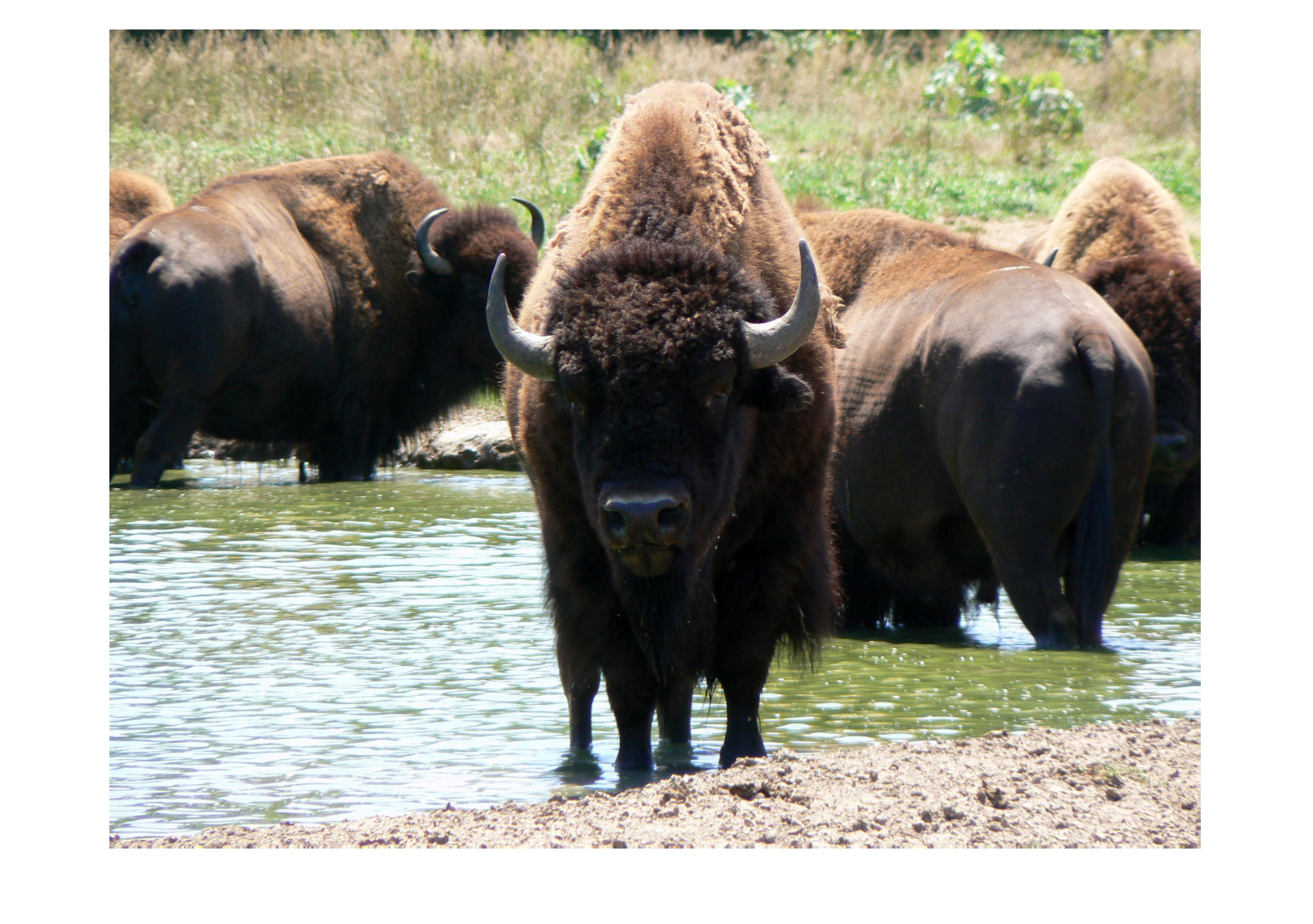}
  \includegraphics[width=0.15\textwidth]{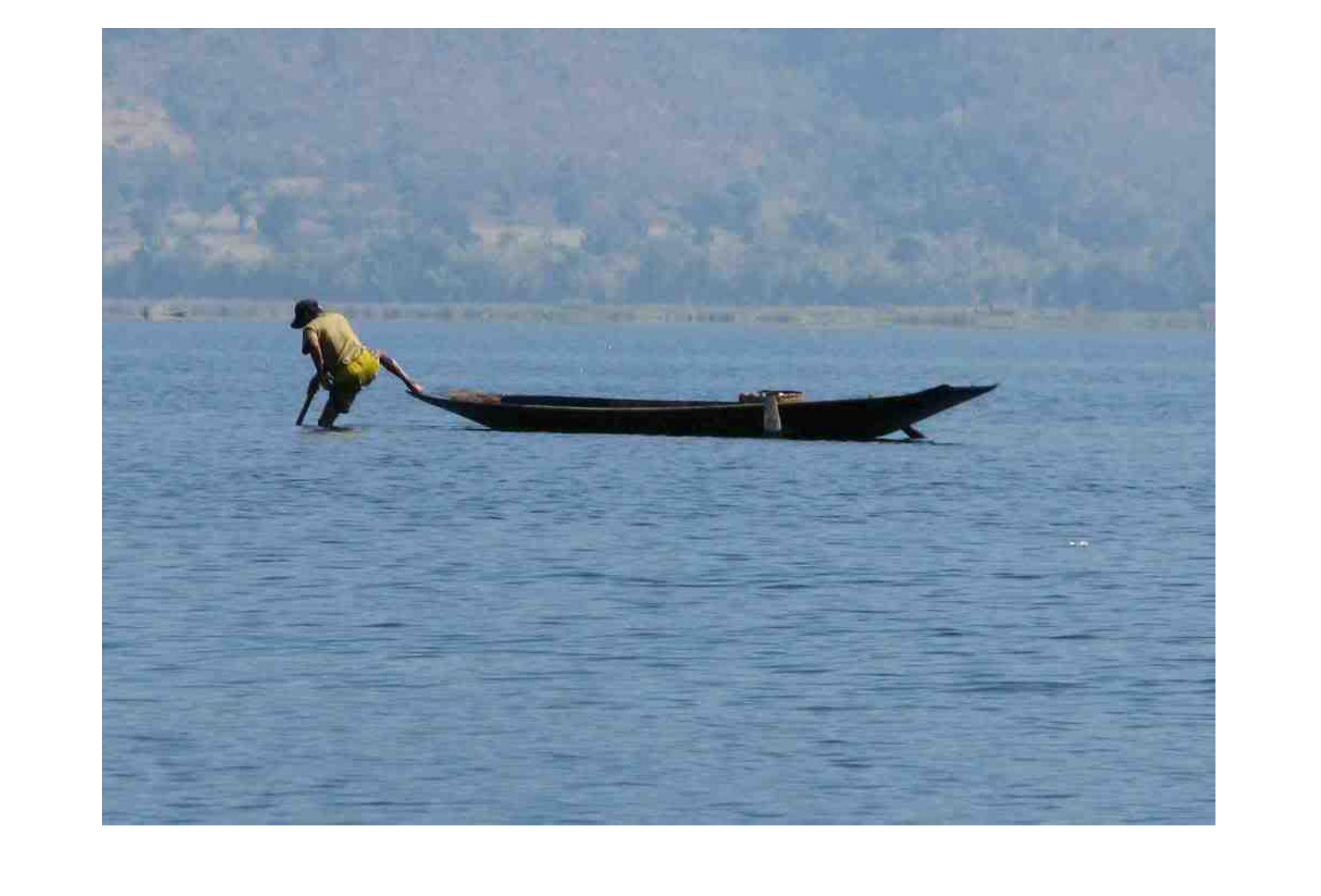}  
  \includegraphics[width=0.15\textwidth]{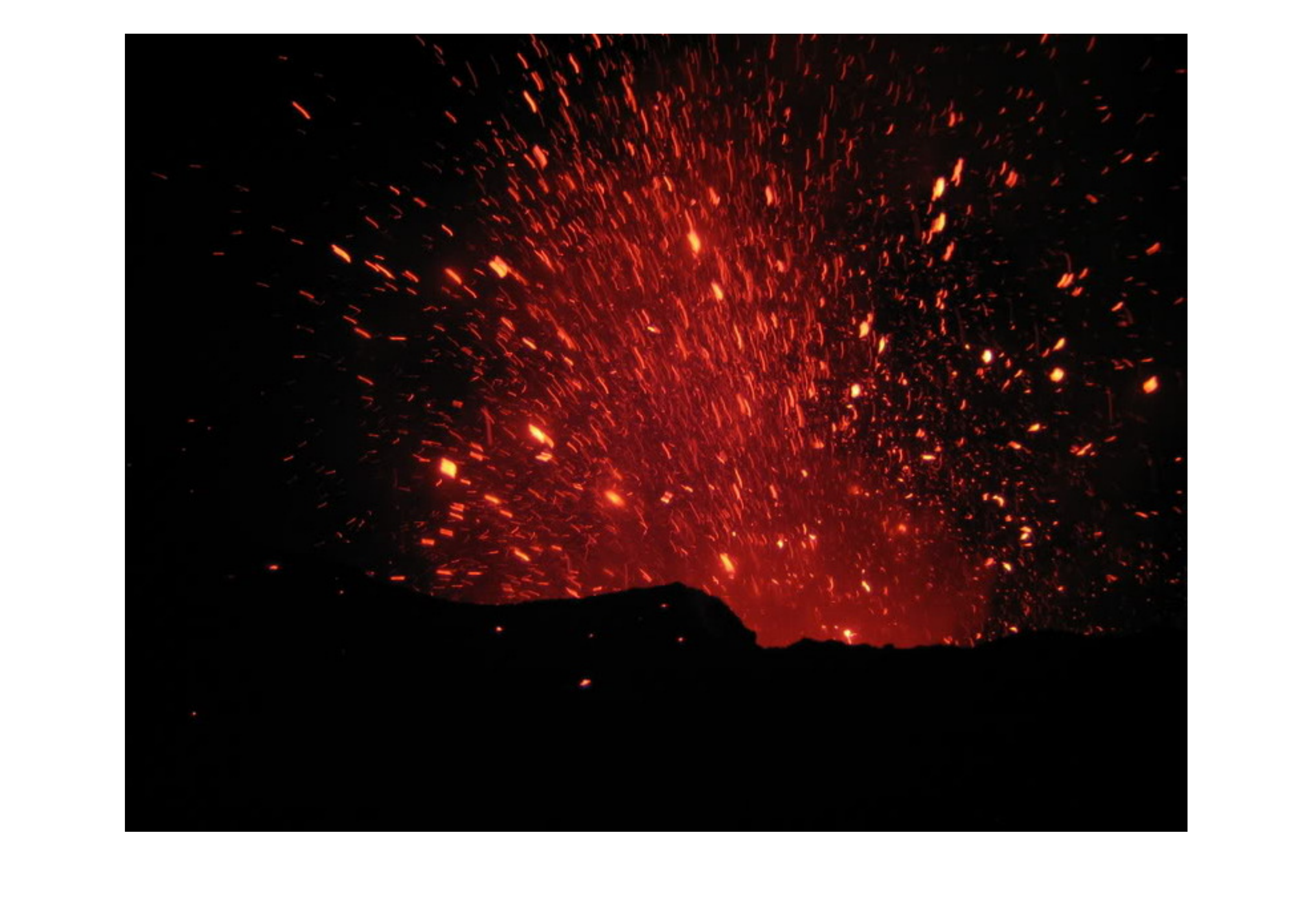}  
  \includegraphics[width=0.15\textwidth]{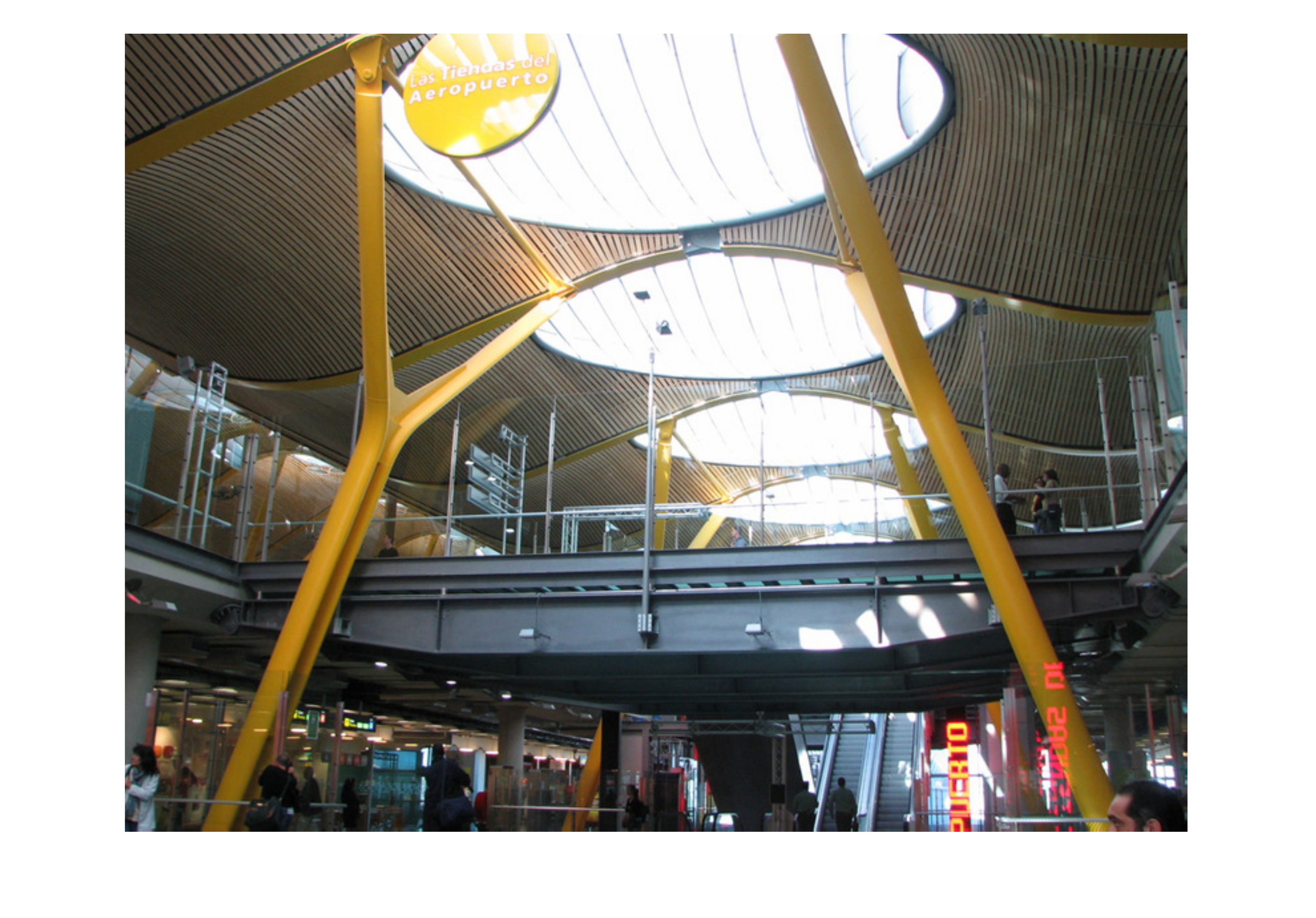}
  \includegraphics[width=0.15\textwidth]{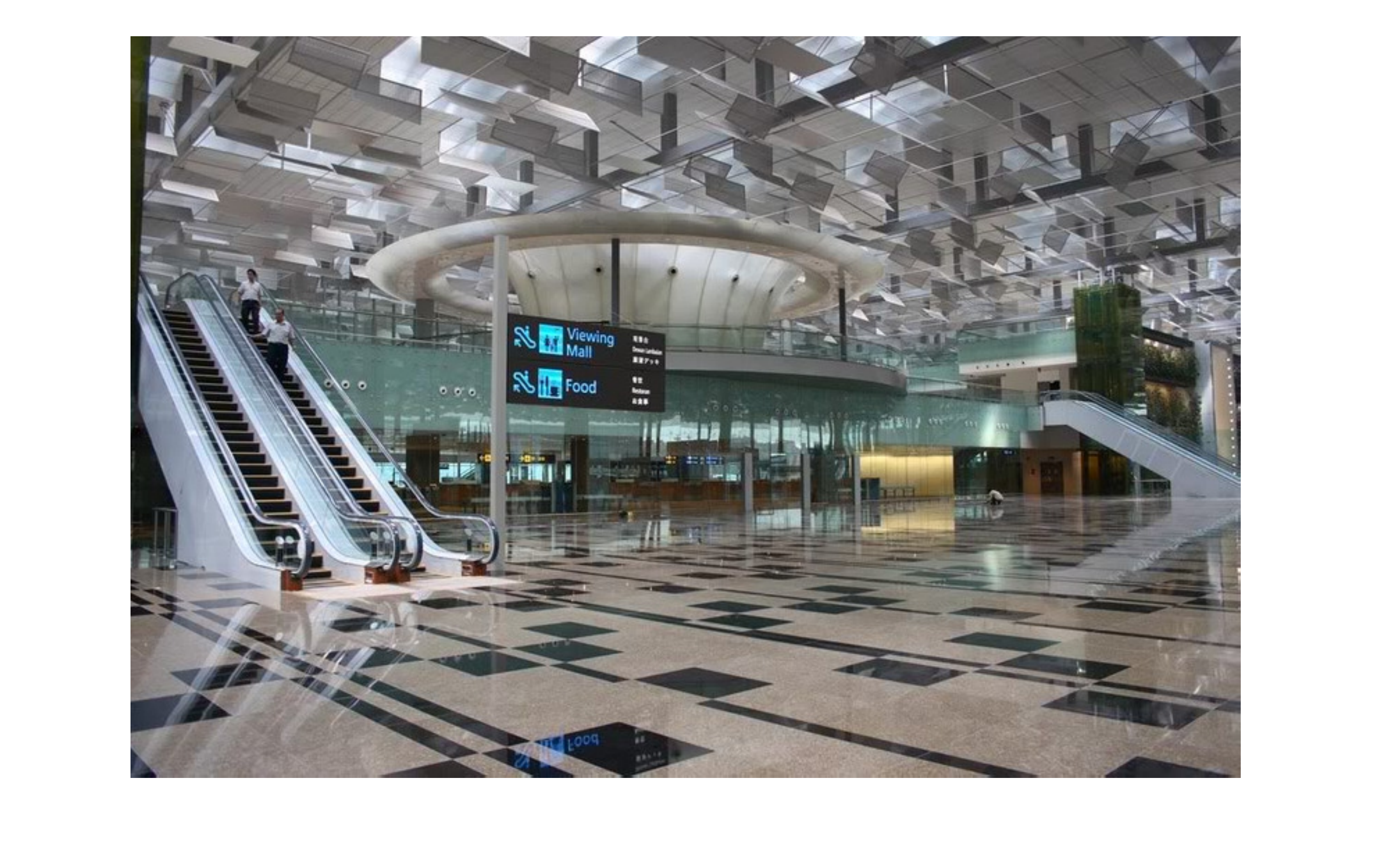}  
  \includegraphics[width=0.15\textwidth]{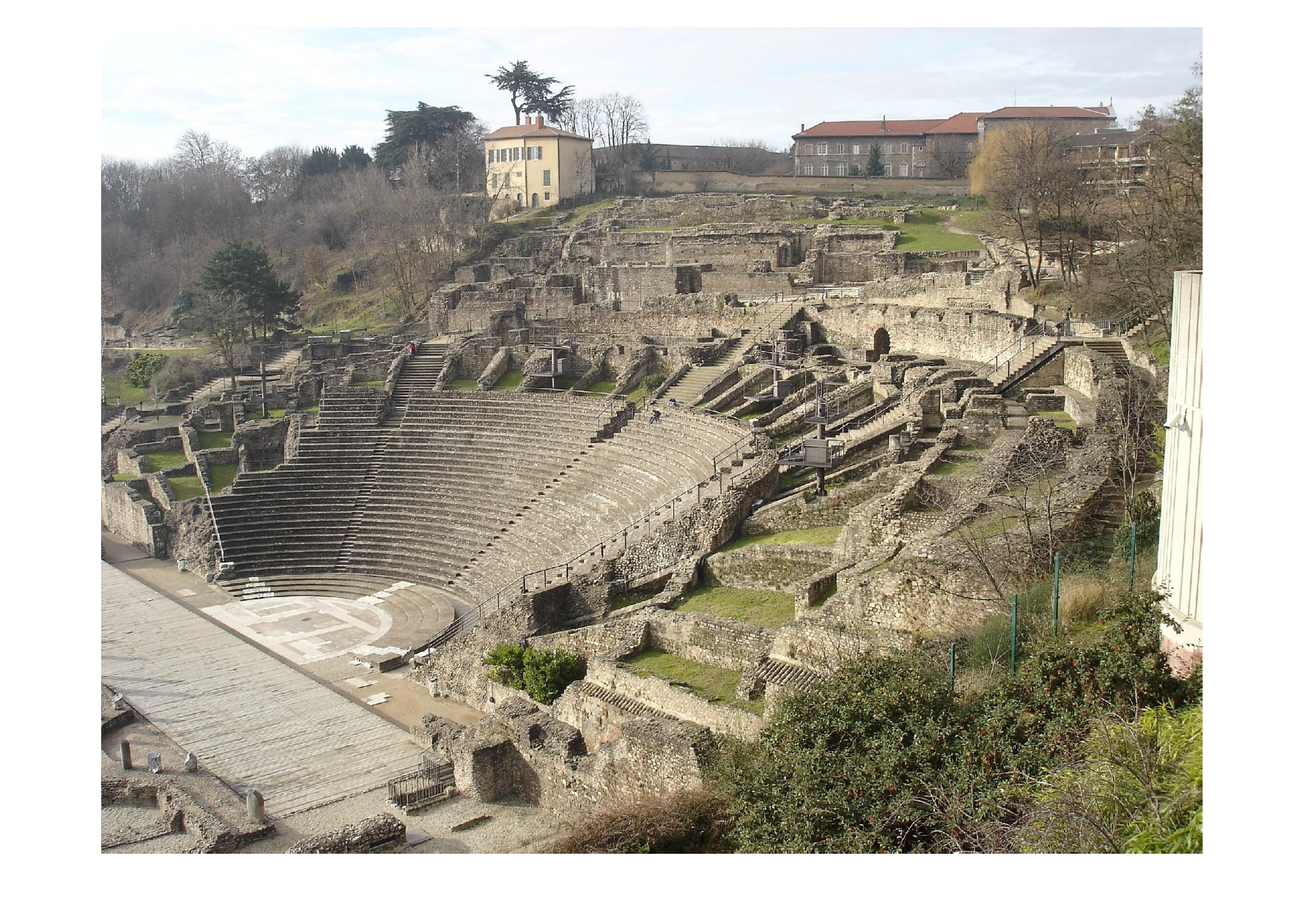}
  \caption{\figlabel{focal_lengths} Example images from the Places dataset from clusters with the largest (up) and smallest (down) focal lengths. Note how images with small focal lengths tend to be more cluttered. A pattern we observed is that dangerous or unaccessible scenes, such as those having volcanos, wild animals and boats tend to be captured using very-high focal lengths,  which is rational.}
\end{figure}

\section{Conclusion}
We have studied the problem of veridical size estimation in complex natural scenes, with the goal of enriching the visual representations inferred by current recognition systems. We presented techniques for performing amodal completion of detected object bounding boxes, which together with geometric cues allow us to recover relative object sizes, and hence achieve a desirable property of any perceptual system - size constancy. We have also introduced and demonstrated a learning-based approach for predicting focal lengths, which can allow for metrically accurate predictions when standard auto-calibration cues or camera metadata are unavailable. We strived for generality by leveraging recognition. This is unavoidable because the size constancy problem is fundamentally ill-posed and can only be dealt with probabilistically.

We also note that while the focus of our work is to enable veridical size prediction in natural scenes, the three components we have introduced to achieve this goal - amodal completion, geometric reasoning with size constancy and focal length prediction are generic and widely applicable. We provided individual evaluations of each of these components, which together with our qualitative results demonstrate the suitability of our techniques towards understanding real world images at a rich and general level, beyond the 2D image plane.
\section*{Acknowledgements}
This work was supported in part by NSF Award IIS-1212798 and ONR MURI-N00014-10-1-0933. Shubham Tulsiani was supported by the Berkeley fellowship and Jo\~{a}o Carreira was supported by the Portuguese Science Foundation, FCT, under grant SFRH/BPD/84194/2012. We gratefully acknowledge NVIDIA corporation for the donation of Tesla GPUs for this research.

{\small
\bibliographystyle{ieee}
\bibliography{main}

\begin{thebibliography}{10}\itemsep=-1pt

\bibitem{baird1963retinal}
J.~Baird.
\newblock Retinal and assumed size cues as determinants of size and distance
  perception.
\newblock {\em Journal of Experimental Psychology}, 1963.

\bibitem{breckon2005amodal}
T.~P. Breckon and R.~B. Fisher.
\newblock Amodal volume completion: 3d visual completion.
\newblock {\em Computer Vision and Image Understanding}, 2005.

\bibitem{burton45}
H.~E. Burton.
\newblock The optics of euclid.
\newblock {\em J. Opt. Soc. Am.}, 1945.

\bibitem{caprile1990using}
B.~Caprile and V.~Torre.
\newblock Using vanishing points for camera calibration.
\newblock {\em International Journal of Computer Vision}, 1990.

\bibitem{imagenet_cvpr09}
J.~Deng, W.~Dong, R.~Socher, L.-J. Li, K.~Li, and L.~Fei-Fei.
\newblock Imagenet: A large-scale hierarchical image database.
\newblock In {\em IEEE Conference on Computer Vision and Pattern Recognition},
  2009.

\bibitem{epstein1963influence}
W.~Epstein.
\newblock The influence of assumed size on apparent distance.
\newblock {\em The American Journal of Psychology}, 1963.

\bibitem{pascal-voc-2012}
M.~Everingham, L.~Van~Gool, C.~K.~I. Williams, J.~Winn, and A.~Zisserman.
\newblock The {PASCAL} {V}isual {O}bject {C}lasses {C}hallenge 2012 {(VOC2012)}
  {R}esults.
\newblock
  http://www.pascal-network.org/challenges/VOC/voc2012/workshop/index.html.

\bibitem{felzens_latent_pami10}
P.~F. Felzenszwalb, R.~B. Girshick, D.~McAllester, and D.~Ramanan.
\newblock Object detection with discriminatively trained part-based models.
\newblock {\em IEEE Trans. on Pattern Analysis and Machine Intelligence}, 2010.

\bibitem{neocognitron}
K.~Fukushima.
\newblock {N}eocognitron: {A} self-organizing neural network model for a
  mechanism of pattern recognition unaffected by shift in position.
\newblock {\em Biological Cybernetics}, 1980.

\bibitem{ghiasi2014parsing}
G.~Ghiasi, Y.~Yang, D.~Ramanan, and C.~C. Fowlkes.
\newblock Parsing occluded people.
\newblock In {\em IEEE Conference on Computer Vision and Pattern Recognition},
  2014.

\bibitem{girshick2013rich}
R.~Girshick, J.~Donahue, T.~Darrell, and J.~Malik.
\newblock Rich feature hierarchies for accurate object detection and semantic
  segmentation.
\newblock In {\em IEEE Conference on Computer Vision and Pattern Recognition},
  2014.

\bibitem{gupta2010blocks}
A.~Gupta, A.~A. Efros, and M.~Hebert.
\newblock Blocks world revisited: Image understanding using qualitative
  geometry and mechanics.
\newblock In {\em European Conference on Computer Vision}, 2010.

\bibitem{Hartley2004}
R.~Hartley and A.~Zisserman.
\newblock {\em Multiple view geometry in computer vision}.
\newblock Cambridge university press, 2003.

\bibitem{hoiem2008putting}
D.~Hoiem, A.~A. Efros, and M.~Hebert.
\newblock Putting objects in perspective.
\newblock {\em International Journal of Computer Vision}, 2008.

\bibitem{Hoiem:book}
D.~Hoiem and S.~Savarese.
\newblock {\em Representations and techniques for 3D object recognition and
  scene interpretation}.
\newblock Morgan \& Claypool Publishers, 2011.

\bibitem{ittelson1951size}
W.~H. Ittelson.
\newblock Size as a cue to distance: Static localization.
\newblock {\em The American Journal of Psychology}, 1951.

\bibitem{kanizsa1979organization}
G.~Kanizsa.
\newblock {\em Organization in vision: Essays on Gestalt perception}.
\newblock Praeger Publishers, 1979.

\bibitem{konkle2011canonical}
T.~Konkle and A.~Oliva.
\newblock Canonical visual size for real-world objects.
\newblock {\em Journal of Experimental Psychology: human perception and
  performance}, 2011.

\bibitem{krizhevsky2012imagenet}
A.~Krizhevsky, I.~Sutskever, and G.~E. Hinton.
\newblock Imagenet classification with deep convolutional neural networks.
\newblock In {\em Advances in Neural Information Processing Systems}, 2012.

\bibitem{lalonde2007photo}
J.-F. Lalonde, D.~Hoiem, A.~A. Efros, C.~Rother, J.~Winn, and A.~Criminisi.
\newblock Photo clip art.
\newblock In {\em ACM Transactions on Graphics (TOG)}, 2007.

\bibitem{LeCun1989}
Y.~LeCun, B.~Boser, J.~Denker, D.~Henderson, R.~E. Howard, W.~Hubbard, and
  L.~D. Jackel.
\newblock Backpropagation applied to hand-written zip code recognition.
\newblock In {\em Neural Computation}, 1989.

\bibitem{palmer1999vision}
S.~E. Palmer.
\newblock {\em Vision science: Photons to phenomenology}.
\newblock MIT press Cambridge, MA, 1999.

\bibitem{russell2009building}
B.~C. Russell and A.~Torralba.
\newblock Building a database of 3d scenes from user annotations.
\newblock In {\em IEEE Conference on Computer Vision and Pattern Recognition},
  2009.

\bibitem{shipley2001fragments}
T.~F. Shipley and P.~J. Kellman.
\newblock {\em From fragments to objects: Segmentation and grouping in vision},
  volume 130.
\newblock Elsevier, 2001.

\bibitem{simonyan2014very}
K.~Simonyan and A.~Zisserman.
\newblock Very deep convolutional networks for large-scale image recognition.
\newblock {\em CoRR}, abs/1409.1556, 2014.

\bibitem{triggs2000bundle}
B.~Triggs, P.~F. McLauchlan, R.~I. Hartley, and A.~W. Fitzgibbon.
\newblock Bundle adjustment -- a modern synthesis.
\newblock In {\em Vision algorithms: theory and practice}. Springer, 2000.

\bibitem{wang1991camera}
L.-L. Wang and W.-H. Tsai.
\newblock Camera calibration by vanishing lines for 3-d computer vision.
\newblock {\em IEEE Trans. on Pattern Analysis and Machine Intelligence}, 1991.

\bibitem{xiang_cvpr15}
Y.~Xiang, W.~Choi, Y.~Lin, and S.~Savarese.
\newblock Data-driven 3d voxel patterns for object category recognition.
\newblock In {\em IEEE Conference on Computer Vision and Pattern Recognition},
  2015.

\bibitem{pascal3d}
Y.~Xiang, R.~Mottaghi, and S.~Savarese.
\newblock Beyond pascal: A benchmark for 3d object detection in the wild.
\newblock In {\em IEEE Winter Conference on Applications of Computer Vision},
  2014.

\bibitem{zhang2000flexible}
Z.~Zhang.
\newblock A flexible new technique for camera calibration.
\newblock {\em IEEE Trans. on Pattern Analysis and Machine Intelligence}, 2000.

\bibitem{zhou2014learning}
B.~Zhou, A.~Lapedriza, J.~Xiao, A.~Torralba, and A.~Oliva.
\newblock Learning deep features for scene recognition using places database.
\newblock In {\em Advances in Neural Information Processing Systems}, 2014.

\bibitem{zia2014towards}
M.~Z. Zia, M.~Stark, and K.~Schindler.
\newblock Towards scene understanding with detailed 3d object representations.
\newblock {\em International Journal of Computer Vision}, 2014.

\end{thebibliography}
}

\end{document}